\documentclass[12pt]{article}
\usepackage[margin=1in]{geometry}

\usepackage{amsmath,amssymb}
\usepackage{cite}
\PassOptionsToPackage{hyphens}{url}\usepackage{hyperref}
\usepackage{xurl}
\usepackage[nopatch=eqnum]{microtype}
\DisableLigatures[f]{encoding = *, family = * }

\usepackage[table]{xcolor}

\usepackage{array, comment}
\makeatletter
\renewcommand{\@biblabel}[1]{\quad#1.}
\makeatother
\usepackage{lastpage,graphicx}

\begin{document}

\title{\Large{Has Sentiment Returned to the Pre-pandemic Level? A Sentiment Analysis Using U.S. College Subreddit Data from 2019 to 2022}}

\author{\normalsize{Tian Yan and Fang Liu}\textsuperscript{*}\\
\normalsize{Applied and Computational Mathematics and Statistics}\\
\normalsize{University of Notre Dame, Notre Dame, IN 46556, United States}\\
\vspace{3pt}
\normalsize{\textsuperscript{*}fliu2@nd.edu}}
\date{}

\maketitle

\begin{abstract}

Background: As the impact of the COVID-19 pandemic winds down, both individuals and society are gradually returning to life and activities before the pandemic. This study aims to explore how people's emotions have changed from the pre-pandemic period during the pandemic to this post-emergency period and whether the sentiment level nowadays has returned to the pre-pandemic level. \vspace{3pt}

Method: We collected Reddit social media data in 2019 (pre-pandemic), 2020 (peak period of the pandemic), 2021, and 2022 (late stages of the pandemic, transitioning period to the post-emergency period) from the subreddits communities in 128 universities/colleges in the U.S., and a set of school-level baseline characteristics such as location, enrollment, graduation rate, selectivity, etc. We predicted two sets of sentiments from a pre-trained Robustly Optimized BERT pre-training approach (RoBERTa) and from a graph attention network (GAT) that leverages both the rich semantic information and the relational information among posted messages and then applied a logistic stacking method to obtain the final sentiment classification. After obtaining the sentiment label for each message, we employed a generalized linear mixed-effects model to estimate the temporal trend in sentiment from 2019 to 2022 and how the school-level factors may affect the sentiment.\vspace{3pt}

Results:  Compared to the year 2019, the odds of negative sentiment in years 2020, 2021, and 2022 are 24\%. 4.3\%, and 10.3\% higher, respectively, which are all statistically significant at the 5\% significance level based on adjusted p-values. In addition, for every 1 standard deviation (18075.6) increase in enrollment, the odds of having negative sentiment in such universities/colleges are 11.9\% higher in a statistically significant manner. Compared to master’s/baccalaureate universities/colleges, the odds of having negative sentiment for doctoral schools with very high research activity is statistically significantly higher with a 30.8\% increase. Region, public vs. private, division I school or not, selectivity, graduation rate, city population, number of doctoral programs, graduate student enrollment, tenured faculty or on tenured track, or having a medical school or not do not affect sentiment in a statistically significant manner in this study. \vspace{3pt} %The results suggest that year change, student enrollment number, and research level of a university are significant factors associated with sentiment change and have a $p$-value $<0.05$.

Conclusions: Our study findings suggest a partial recovery in the sentiment composition (negative vs. non-negative) in the post-pandemic-emergency era.  The results align with common expectations and provide a detailed quantification of how sentiments have evolved from 2019 to 2022 in the sub-population represented by the sample examined in this study. \vspace{3pt}%We have also identified various factors that influence changes in sentiments. 

\end{abstract}

\textbf{Keywords}: COVID; pandemic; post-pandemic; sentiment temporal trend; Reddit data; social media; machine learning; RoBERTa, graph attention mechanism; Generalized linear mixed models; negative; positive

\section{Introduction}\label{sec:intro}
\subsection{Background}\label{sec:background}

% and the  Center for Disease Control and Prevention (CDC) mask mandate for airplanes and public transport was struck down as unlawful.
While COVID-19 remains a public health priority, many governments have transitioned away from the emergency phase that gripped the globe in 2020 and 2021. With a variety of effective strategies implemented to combat the COVID-19 pandemic, including vaccination, quarantine measures, and the adoption of remote work and study routines, the impact of the pandemic on society has gradually subsided since the second half of 2021. In the U.S., nearly all state-level mask mandates had been lifted by April 2022; many educational institutions from elementary schools to higher education institutes have returned to the pre-pandemic in-person learning mode; social gatherings, conferences, sports, and entertainment events have also welcomed back participants and fans at full capacity, among others. 
%The U.S. Department of Health and Human Services declared the COVID-19 public health emergency officially over as of May 11, 2023. 

However, the post-pandemic world does not mirror the pre-pandemic era in many aspects, including the psychological and emotional aspects. % In the context of higher education, universities have made divergent decisions, with some opting to continue offering online learning, while others have mandated the resumption of in-person classes. Additionally, there exists a spectrum of attitudes and sentiments among individuals regarding the reopening of the economy and the restoration of normalcy. \cite{rahman2020twitter}
Several studies have been conducted to analyze sentiments and attitudes on various aspects post the COVID-19 pandemic. 
\cite{bustos2022twitter} studied Twitter users' sentiment change toward COVID-19 vaccination after the first COVID-19 vaccination was implemented in the U.S.. They used negative binomial regression and linear regression and found that public sentiment towards vaccination became more positive after the first dose of vaccination.  Through frequency verification between public opinions and sentiments and analysis of the influence mechanism, 
\cite{yi2022appeal} claimed positive public opinion and sentiment on ports and corporate choice of import and export of goods post the pandemic.  \cite{chekijian2021emergency} studied sentiment and topic trends related to patient experience before, during, and after the pandemic in the framework of topic modeling with latent Dirichlet allocation.  %and found that people most cared about technical competence in the pre-pandemic period, and the concerns shifted to appointment activity and COVID-19 related topic during pandemic, and cared aboout interpersonal manners most post pandemic. 
\cite{rahman2020twitter} studied the socioeconomic factors that may affect people's attitude towards reopening the economy in post-COVID-19 using Twitter data, socioeconomic data, environment data, and COVID-19 cases, and applied logistic regression to identify important factors. They concluded that people with low education levels, low income, in the labor force, and with higher residential rents are more interested in reopening the economy. 
\cite{ismail2022triggers} studied the public well-being and sentiment toward education post-COVID-19. They curated Twitter data that's relevant to the education sector and used Aspect-based Sentiment Analysis and machine learning techniques to identify sentiment and emotional triggers. They conclude that safety is a top concern for students, parents, and educators. 
\cite{qaqish2023sentiment} studied the Jordanian community’s attitude towards online and in-person hybrid learning by analyzing post-pandemic Twitter data. They used long short-term memory (LSTM) networks to classify tweet emotions and found that 18.75\% of the samples fall within the category of Dissatisfied Anger and Hate, 21.25\%  Sad, 13\%  Happy, and 24.5\%  Neutral. 
\cite{bhalla2021spiritual} studied the motivation and inclination of traveling in 2021 using thematic analysis, sentiment classification, and word cloud. They concluded that nature-based travel has become the first choice of travel after 2020. 
\cite{saura2022exploring} used Twitter data to study people's attitudes towards remote working in the post-pandemic era. They used TextBlog, the latent Dirichlet allocation  model, and multiple machine learning models, and found that the topics ``work-life balance'', ``less stress'',  ``future'' and ``engagement'' are positive; negative topics include ``virtual health'', ``privacy concerns'', and ``stress'', and neutral topics involve ``new technologies'', ``sustainability'', and ``technology issues''.
\cite{shopnil2022post} analyzed Twitter data to study the sentiment distribution in India after the second wave of COVID-19 using Vader, LSTM, and convolutional neural networks and found the majority of sentiments are either neutral or positive. 
%\cite{dasgupta2022sentiment} studied consumers' attitudes toward car buying during and after the lockdown. They scrapped the data from Twitter, YouTube, and online news, and concluded that facelifts, spot tests, voice bots, discounts, advertising, and better financing options are key factors of reviving car buying.

In summary, all the above work used social media data to study sentiment or attitudes at a single pandemic time point in 2021 and 2022 after the first wave of the pandemic in 2020, with many focusing on data in a specific domain such as travel, import/export, remote working, and education.

\subsection{Study Objective and Overview}
This study is different from the works summarized in Section \ref{sec:background}. First, it investigates general sentiment rather than sentiment in a specific domain or attitude toward a specific topic; in addition, it examines the temporal trend of sentiment from 2019 to 2022, representing the before-pandemic baseline and several phases during the pandemic, rather than a snapshot in time.

This study is a follow-up study to \cite{yan2022covid} that examined the sentiment during the early phase of the pandemic (2020) as opposed to the pre-pandemic (2019) in 8 higher-education institutes (HEI) with Reddit data in the U.S. In this study, we collected Reddit data from 128 HEIs in the U.S., including the 8 schools in \cite{yan2022covid}, over a 4-year period (August to December in 2019, 2020, 2021, and 2022), where 2021 and 2022 can be regarded as later stages of the pandemic. In other words, the scope of this study is much broader with a longer study period and many more schools that cover all four regions of the U.S.; the number of messages also increases from 165,570 in \cite{yan2022covid} to 4,129,170 in this study.

The primary goal of this study is to examine the sentiment shift from 2019 to 2022 and whether and when the level of negative sentiments has returned to the pre-pandemic era (2019).  As secondary objectives, we also examine how other factors may affect the sentiment based on the collected Reddit data, such as region, school type and classification, enrollment, etc. 

To analyze the data, we adopted a similar approach as in \cite{yan2022covid} by first predicting the sentiment of each collected message using machine learning. The technique employs advanced natural language processing (NLP) techniques, specifically, the Robustly Optimized BERT Pretraining Approach (RoBERTa), in conjunction with graph neural networks (GNN) that leverage the inter-message relations among the Reddit messages %Our primary objective was to analyze the evolving sentiments within a dataset encompassing 128 higher education institutes (HEIs) situated in the U.S. over distinct temporal phases: pre-COVID-19, during the pandemic, and post-COVID-19. Subsequently, we integrated the RoBERTa and GNN models through a logistic stacking approach, leveraging the combined strength of these techniques to construct our sentiment classifier.  This integrative approach allowed us to capitalize on both the rich semantic content and the 
Upon assigning sentiment classes (negative or non-negative) to each message, we employed a generalized linear mixture model (GLMM) to examine the effect of year on sentiment and to identify relevant covariates that may have significant relations with sentiment. %proceeded to visualize the distribution of negative message percentages for each respective year. This comprehensive analysis enabled us to discern trends and patterns in sentiment shifts across various temporal phases. Subsequently, we treated the sentiments of individual messages as response variables, with year, HEI type, graduation rates, and other HEI-specific attributes serving as predictor variables. These variables were employed in a 

The remainder of the paper is structured as follows. In Section \ref{sec:data}, we describe the data collection for this study. In Section \ref{sec: methods}, we introduce the machine learning and statistical procedures used to make sense of the data. The study results are presented in Section \ref{sec:results}. The study limitations and future work are discussed in Section \ref{sec:discussion} and the main study conclusions are described in Section \ref{sec:conclusion}.

%------------------------------------------------
\section{Data Collection}\label{sec:data}
Our study focuses on a sample of Higher Education Institutions (HEIs) in the U.S. with Reddit data. When selecting the school, we aimed for representativeness and diversity. We first compiled a list of HEIs with subreddits, leading to more than 400 institutions. We then dropped those schools that don't have enough messages in their subreddits. Specifically, if a subreddit has $<20 $ messages in each of the four years from 2019 to 2022 or has at least two years with $<10$ messages, we dropped the school from the sample. In addition, due to storage and computational constraints (see Sec \ref{sec: methods}), we subsetted the schools, eventually leading to a total of 128 schools, as listed in provided in the Appendix. When subsetting the schools, we used criteria such as school diversity in terms of geographical regions within the U.S. and HEI types (i.e., research universities, liberal arts colleges, and institutions specializing in particular fields such as the Naval Academy). The schools also have a wide of range rankings from more prestigious institutions to colleges/ universities ranked beyond 300 per the U.S. News rankings. Nevertheless, it's important to acknowledge that the selection process, to some extent, involved subjective judgment influenced by the authors' knowledge, despite our best efforts to maintain objectivity.

The data collection process started with the retrieval of all textual messages from the subreddits associated with each HEI. The time frame spanned from August to November in each year from 2019 to 2022. We supplemented this textual corpus with additional attributes specific to each HEI, such as region, Carnegie classification of HEI (CCHEI), enrollment,  graduation rate,  faculty headcount, etc.

\subsection{Reddit data}\label{sec: reddit}
The Reddit data collection and how the data are used in this study are in accordance with Reddit’s Terms and Conditions on data collection and usage. We also consulted the research compliance program at the University of Notre Dame and no IRB approval is needed given that the collected data are publicly accessible on Reddit; we did not collect private identifiable data nor interacted with the Reddit users. More information regarding privacy compliance is provided in the Appendix.

To examine the sentiment trend from 2019 and 2022, we downloaded the data from August to November, in the year 2019, 2020, 2021, and 2022 from the subreddit communities of the 128 schools.  2019 is regarded as the pre-pandemic baseline, 2020 was during the pandemic, while 2021 and 2022 represent the transition to the post-emergency period. We used the Pushshift API (https://github.com/pushshift/api) to download the comment data but excluded the submission data due to the non-availability of the submission data when the study was conducted.

The messages in each school form a graph. In the graph, each message represents a node, and if one message replies to another one, they are direct neighbors and we draw a directional edge between the two. Due to computational constraints, we limited the size of each graph. For those schools with over 30,000 nodes, we obtain a subgroup with 30,000 nodes using the following sampling approach. We first randomly select 50 nodes to start and then add the nodes that are connected to at least one of the 50 nodes one by one to form the subgraph. If there are no direct neighbors to any of the 50 nodes, we randomly select another 50 nodes to add to the subgraph. When the subgraph node number gets close to 30,000 in the node-adding process, we only add the first several neighbor nodes or the first several randomly selected nodes to make it exactly 30,000. In this case, the sequence of the neighbor nodes based on which they will be added is determined by a combination of the Breadth First Search algorithm\cite{moore1959shortest} and the sequence that the neighbor message appears in our data (an index that is independent of the messages themselves). We repeat this process until the subgroup meets the threshold of 30,000 nodes.

\subsection{Baseline school-level data}\label{sec: meta}
We considered a set of variables at the school level that might impact the sentiment change from 2019 to 2022. % Generally speaking, academic requirements, economic condition, and climate\cite{hannak2012tweetin} are several important factors that could potentially influence a student's sentiment, thus we selected those variables that could reflect the impact of these factors. 
The baseline data were collected from multiple sources -- the 2020 United States census\cite{CENSUS}, Carnegie Classification Of Institutions Of Higher Education (CCIHE)\cite{CC}, and  Integrated Postsecondary Education Data System (IPEDS)\cite{IPEDS}. The variables are listed in Table \ref{tab:meta}. %2020 United States Census, Type and Admission are from CCIHE, and the rest of them are from IPEDS. 
For the data collected from IPEDS, if there are multiple years of data, we average them across all available years to obtain the final value for these variables. 

%----------------------------------------------
\section{Methods}\label{sec: methods}
We apply several machine learning and statistical analysis approaches to explore the data and address the goal of this study. 

\subsection{Sentiment Prediction}\label{sec: gat}
We apply the same ensembled graph neural networks and pre-trained RoBERT model in \cite{yan2022covid} to predict the sentiment class (negative vs. nonnegative) for each message in the downloaded Reddit dataset. 

RoBERTa\cite{liu2019roberta} is an improved version of the BERT (Bidirectional Encoder Representations from Transformers)\cite{devlin2018bert} model and a pretraining framework that's based on the attention mechanism\cite{vaswani2017attention}. The original RoBERTa was trained on a dataset of over 160GB of uncompressed text, and it includes BookCorpus plus English Wikipedia (16GB), CC-News (76GB), OpenWebText (38GB) Stories (31GB). However, the Reddit data we used has different properties than the original training data. It contains emojis, non-standard spelling, Internet slang, and other possible features of Internet language. Thus we choose a RoBERTa model\cite{barbieri2020tweeteval} that is trained on ~58 million messages from Twitter and fine-tuned for sentiment analysis, which is more suitable for our application. The  Python  code  for  the  RoBERTa framework is  adapted  from \cite{barbieri2020tweeteval} (see the Appendix for the link to the code).

We obtained embeddings for the Reddit messages from the RoBERTa model that are used in two downstream learning tasks. First, the embeddings are fed to a feed-forward neural network with softmax as the last layer to output the sentiment probabilities for the messages. Second,  to better utilize the relational information among messages, we employ the Graph Attention Networks (GAT) model trained in \cite{yan2022covid} with the embeddings and the adjacency matrices among the messages as input to output a second set of predicted sentiment probabilities for the messages. GAT is a kind of GNN model that incorporates the attention mechanism into the graph. In our case, we treat all messages in each school as an independent graph. When one message replies to another, an edge goes from the first message to the second message but not the other way around. In other words, the adjacency matrix is asymmetric. The GAT model updates the hidden state of each node given the initial states of the node and its neighbors.

\cite{yan2022covid} found that GAT and RoBERTa can be inconsistent in their sentiment prediction -- GAT tends to be more accurate in predicting negative messages, and Roberta tends to be more accurate in non-negative messages. 
To obtain more accurate sentiment predictions, we applied the stacking method in \cite{yan2022covid} and formulated a logistic model to combine the sentiment probabilities from GAT And RoBERTa to obtain the final sentiment classification for each message. 

Regarding the computational cost for running the prediction models, it took about one week to run RoBERTa and one day to run GAT, respectively, across all the messages in 128 schools on a computer with Intel(R) Xeon(R) CPU L5520  @ 2.27GHz and RAM 72.0 GB, and x64-based processor. 98.7 GB was used to store all unprocessed and processed data.

\subsection{Statistical Analysis of Sentiment Trend from 2019 to 2022}\label{sec: glmm}
After having the sentiment classifications for the 4,129,170 messages, we fitted a generalized linear mixed-effects model (GLMM) to examine how sentiment changes from 2019 and 2022. %August to December 2019  represents the pre-pandemic era, August to December 2020 and 2021 represent the pandemic period, and August to December 2022 presents the post-pandemic era. 

The GLMM is $\text{log}\left({\frac{\Pr(y_{ik} \text{ is negative})}{1-\Pr(y_{ik}\text{ is negative})}}\right)=\beta_0+\sum_{j=1}^p\beta_j x_{ijk}+z_k$, where the sentiment label  $y_{ik}$ (negative vs non-negative) of message $i$ in school $k$ is the binary response;  year (categorical) and the set of variables in Table \ref{tab:meta} are fixed-effect predictors coded in $x_{ijk}$ for $j=1,\ldots,p$ ($p$ is the number of regression coefficients associated with covariates $\mathbf{X}$). Because the messages from the same school are correlated,  $z_k\sim \mathcal{N}(0,\sigma^2)$ is included as a random effect to account for the within-school dependency.

%---------------------------------------------------------
\section{Results}\label{sec:results}
\subsection{School-level Characteristics}
The baseline characteristics of the school-level data are summarized in Tab \ref{tab:meta}. For categorical variables, frequency and percentage of each category are provided; for continuous variables, mean, standard deviation, minimum, and maximum are provided.
\begin{table}[!htb]
\centering 
\caption{ Descriptive Statistics of School-level Baseline Characteristics}\label{tab:meta}\vspace{3pt}
\resizebox{1\textwidth}{!}{
\begin{tabular}{lcc} 
\hline
Variable & summary statistics & Source and Year\\
\hline
\multicolumn{3}{c}{categorical variable: count (percentage) }\\
\hline

Region & & - -\\%& - \\
\:\:\:\:West& 24 (18.75\%) \\
\:\:\:\:South & 41 (32.03\%)\\
\:\:\:\:Northeast & 31 (24.22\%) \\
\:\:\:\:Midwest & 32 (25.00\%) \\

Type & & CCHIE, 2021 \\ 
\:\:\:\:Private& 44 (34.38\%) \\
\:\:\:\:public& 84 (65.63\%) \\

D1 (NCAA Division 1 school) & & NCAA, 2023 \\
\:\:\:\:Yes& 108 (84.38\%) \\
\:\:\:\:No& 20 (15.62\%) \\

CCHIE (Carnegie classification)& &  CCHIE; 2021 \\
\:\:\:\:Baccalaureate: arts \& sciences focus& 7 (5.47\%) \\
\:\:\:\:Master's: larger programs& 1 (0.78\%) \\
\:\:\:\:Doctoral: high research activity& 18 (14.06\%) \\
\:\:\:\:Doctoral: very high research activity& 102 (79.69\%) \\

Medical (grants a medical degree?) &   & IPEDS; 2021 \\ 
\:\:\:\:Yes& 77 (60.16\%) \\%&(MD, DDS, DMD, DO, DVM) \\
\:\:\:\:No& 51 (39.84\%) \\
\hline
\multicolumn{3}{c}{continuous variable: mean $\pm$ SD (min, max) }\\
\hline
Population (city population in 1,000)& 564.4 $\pm$ 1278.8 (7.2, 8804.2)& U.S. 2020 Census \\%& 231 \\

Doctoral program (No. of doctoral programs) & 315.7 $\pm$ 238.7 (0, 876) & IPEDS; 2021 \\

Tenure (tenured/on tenure track faculty count) & 1152.3 $\pm$ 682.9 (150, 3280)  & IPEDS; 2019 - 2021 \\

Enrollment (12-month unduplicated  enrollment \\
\hspace{3pt} in 1,000)& 30.8 $\pm$ 18.1 (1.5, 101.9) & IPEDS; 2019 - 2021 \\%& 231 \\

Graduate student (12-month unduplicated  \\
\hspace{3pt} graduate student enrollment  in 1,000) & 9.020 $\pm$ 6.153 (0, 2.920)  & IPEDS; 2020 - 2021 \\ 

Selectivity (Percent of applicants admitted) & 0.525 $\pm$ 0.291 (0.050, 0.964)  & IPEDS; 2020\\ 

Rate (graduation rate of bachelor\\
\hspace{3pt} degree within 4 years) & 61.2 $\pm$ 20.8 (15.7, 91.3) & IPEDS; 2019 - 2021 \\%& 231 \\
\hline
\end{tabular}}
\end{table}

\begin{figure}[!htb]
\includegraphics[scale=0.75]{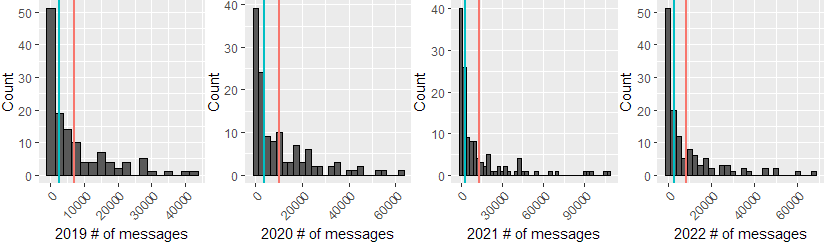}
\caption{ Histograms of number of messages across schools by year (blue and red lines represent median and mean, respectively)}\label{fig:messages}
\end{figure}

Fig \ref{fig:messages} depicts the distributions of the number of messages across the 128 schools by year. In each year, the number of messages varies by school, but most schools have messages $<30k$ across all 4 years. Due to computational constraints, for schools with more than 30k messages, we sampled a subgraph that has 30,000 messages (node) using the methods described in Section \ref{sec: reddit}. This leads to a total of 4,129,170 messages the sentiment of which are predicted. %Message numbers increase over year, and the mean of them gradually increased from around 7,000 in 2019, to around 16,000 in 2022.

\subsection{Sentiment classification}
\begin{figure}[!htb]
\includegraphics[scale=0.85]{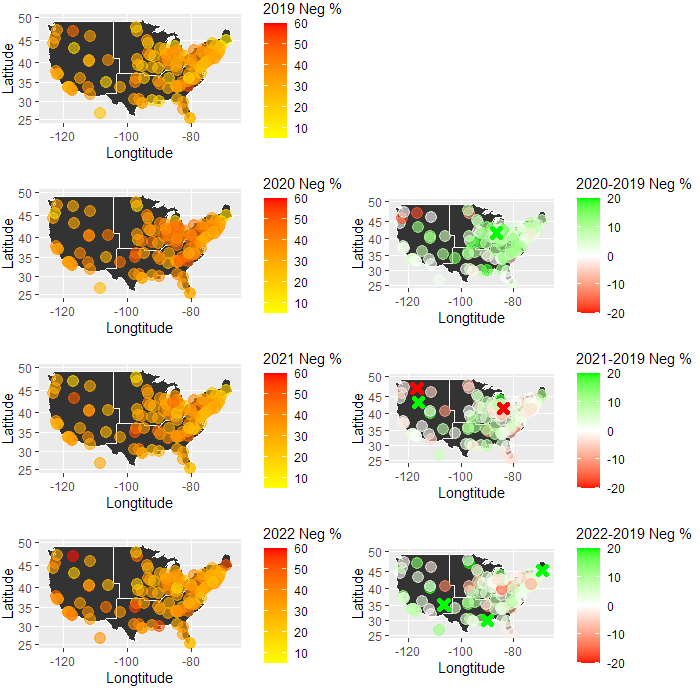}
\caption{Heatmaps of negative percentage in all schools by year. Each circle represents a school. The right column shows the within-school differences in  2020 to 2022 vs. 2019 (pre-pandemic). The crosses in the 2021 and 2022 plots represent difference values outside the $[-20,20]\%$ range (28.04\% for the University of Notre Dame in 2020; 20.01\% for Bowling Green State University-Main Campus,32.35\% for
Boise State University, and -27.83\% for the University of Idaho in 2021; and 44.30\% for the University of Maine, 26.18\% for the 
University of New Mexico-Main Campus, 32.34\% for the 
Tulane University of Louisiana in 2022)}\label{fig:map}
\end{figure}

\begin{figure}[!htb]
\includegraphics[scale=0.9]{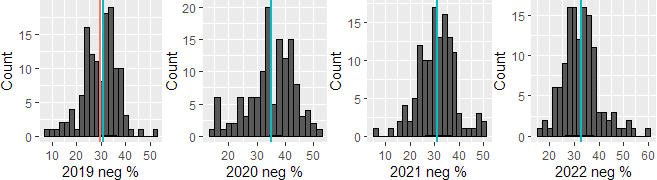}
\caption{Percentage of Negative sentiment distribution for all schools in each year (blue and red  line represents median and mean, respectively)}
\label{fig:negperc}
\end{figure}

We calculate the percentage of negative messages in each school year based on the predicted sentiment from the stacked model and present the heatmap results (left column) in Fig \ref{fig:map}. In the maps, each circle represents one school and its position in the map represents its geographical location in the U.S. In addition, we also plot the difference in the percentage of negative sentiment from 2020 to 2022 vs. 2019 for each school (right column) in Fig \ref{fig:map}. Compared to the year 2019, all the other years have higher negative sentiment proportions than 2019. The year 2020 has the highest negative percentage. For years 2021 and 2022, although they have higher negative sentiment proportions than 2019, they are still more positive than the year 2020. Similar trends can be observed in Fig \ref{fig:negperc} that depicts the distribution of negative sentiment percentage across the schools by year. 

\subsection{Temporal sentiment trend from 2019 to 2022 and school-level covariate effects on sentiment}
\begin{figure}[!htb]
\includegraphics[scale=1, trim= 16pt 6pt 6pt 24pt, clip]{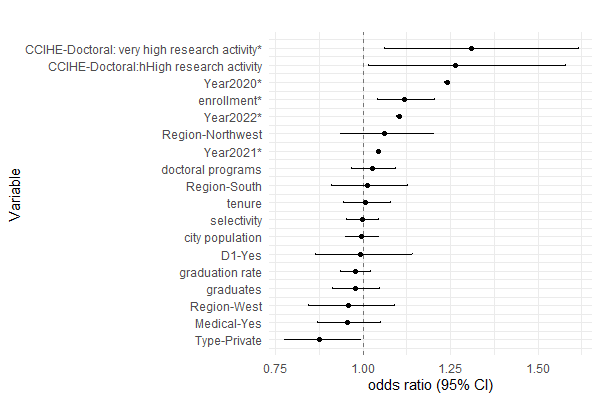}
\caption{Forrest plot of estimated odds ratios of negative sentiment with 95\% confidence intervals. An asterisk * indicates a statistically significant odds ratio per the adjusted p-value (Table \ref{tab:glmm}) for the corresponding covariate vs. its reference level or with 1SD increase in the variable} \label{fig:forest}
\end{figure}

The GLMM model was run on complete records only (a total of 4,129,170 messages). 
%We assume the missingness is missing at random or completely at random. In addition, given the very low missing fractions (Table \ref{tab:meta}, we expect running the model on only complete record would not impact the results, statistical inference, and the final conclusions, even if some missing values were subject to not-at-random missingness.  
There is a high imbalance in the covariate CCHIE, with only one school classified as ``Master's Colleges \& Universities: Larger Programs'' and seven schools as ``Baccalaureate: arts \& sciences focus'', which could lead to potential computational and inferential problems in the GLMM estimation.  We thus combined the two categories as one and referred to it as ``Baccalaureate/Master's Colleges/Universities''. The inferential results from the GLMM are presented in Table \ref{tab:glmm} and Fig \ref{fig:forest}. We used the \texttt{glmer} function in R package \emph{lme4} to run the GLMM and the \texttt{p.adjust} function in R package \emph{stats} to obtain FDR  adjusted $p$-values.  

Using $<0.05$ as the threshold for statistical significance on the adjusted p-values, Year is significantly associated with Sentiment. The odds of having negative sentiments in 2020,  2021, and 2022, are 24\%, 4.3\%, and 10.3\% higher, respectively, than that in 2019, suggesting the likelihood of negative sentiment increased significantly during the pandemic compared to before the pandemic, based on the messages posted on Reddit, the negative sentiment proportion in the later half of 2021 almost returned to the pre-pandemic level, and slightly increased during the second half of 2022.  Overall, we may conclude the level of negative sentiment goes down post-pandemic compared to during the pandemic, not is still higher than pre-pandemic. 

Enrollment is also statistically significantly associated with sentiment. For every one SD (18,075) increase in enrollment, the odds of having negative sentiment goes up by 11.9\%, implying larger enrollment tends to have a negative impact on Sentiment. Compared to Master's/Baccalaureate universities/colleges, the odds of having negative sentiment for doctoral schools with very high research activity is 30.8\% higher. Though the odds for doctoral school with high research activity is also higher (26.5\%) than that for Master's/Baccalaureate universities/colleges, the increase is not statistically significant at the 5\% level based on the adjusted p-value. The observations exhibit rational comprehensibility; there is constant pressure on both students and faculty members in HEIs with the requirement of research productivity and excellence, which is likely linked with the higher negative sentiment in those schools and the communities that are associated with them.

Private schools tend to have lower odds of negative sentiment  (12.4\% lower) than public schools, though the difference is not statistically significant based on the adjusted $p$-value. The rest of the examined covariates do not have a pronounced effect on Sentiment, such as region, D1 school or not, a medical school or not, selectivity, etc.

The CI widths for the odds ratios on the year comparisons are much smaller compared to those associated with other factors. This is because Year is the only within-cluster factor whereas the others are between-cluster factors, where cluster here refers to School in the mixed-effect model. The variance of the effect of a between-cluster factor contains the between-cluster variance (variance across schools) and the sampling variability whereas that for a within-factor factor only contains the latter and is thus smaller. The very precise estimates for the year comparison benefit from the huge number of messages, where the precision on the estimated effects of the between-school factors is more determined by the number of schools, which is 128.

\begin{table}[!htb]
\caption{Estimated effects of covariates on the odds of negative sentiment} \label{tab:glmm}\vspace{-3pt}
\begin{center}
\resizebox{0.95\textwidth}{!}{\begin{tabular}{l@{\hspace{3pt}}l@{\hspace{6pt}}c@{\hspace{3pt}}c@{\hspace{0pt}}c} 
\hline
&&odds ratio & \multicolumn{2}{c}{ $p$-value } \\
\cline{4-5}
\multicolumn{2}{l}{Covariate} & (95\% CI) & raw & adjusted$^\dagger$\\
\hline
Region & Midwest & - & - & -\\
& Northwest & 1.060 (0.936,1.201) & 0.361 & 0.623\\
& South & 1.013 (0.910,1.127) & 0.817 & 0.919\\
& West & 0.958 (0.842,1.091) & 0.520 & 0.706\\
\hline
Type & Public &  - & - & -\\
& Private & 0.876 (0.774,0.991) & 0.035 & 0.088\\
\hline
Year & 2019 &  - & - & -\\
& \textbf{2020} & \textbf{1.240 (1.233,1.248)} & $\mathbf{<\!0.001}$ & $\mathbf{<\!0.001}$\\
& \textbf{2021} & \textbf{1.043 (1.037,1.049)} &$\mathbf{<\!0.001}$ & $\mathbf{<\!0.001}$\\
& \textbf{2022} & \textbf{1.103 (1.096,1.109)} & $\mathbf{<\!0.001}$& $\mathbf{<\!0.001}$ \\
\hline

D1& No &  - & - & -\\
& Yes & 0.993 (0.863,1.142) & 0.919 & 0.919\\
\hline
CCHIE & Baccalaureate or Master's &  - & - & -\\
& Doctoral: high research activity & 1.265 (1.014,1.578) & 0.037 & 0.088\\
& \textbf{Doctoral: very high research activity} & \textbf{1.308 (1.061,1.614)} & \textbf{0.012} & \textbf{0.038}\\
\hline
Medical & No  &  - & - & -\\
& Yes & 0.955 (0.869,1.05) & 0.343 & 0.623\\
\hline
\multicolumn{2}{l}{city population$^\ddag$} & 0.996 (0.95,1.044) & 0.875 & 0.919\\
\multicolumn{2}{l}{\textbf{enrollment$^\ddag$}} & \textbf{1.119 (1.041,1.203)} & \textbf{0.002} & \textbf{0.009}\\
\multicolumn{2}{l}{doctoral programs$^\ddag$} & 1.027 (0.966,1.092) & 0.394 & 0.624\\
\multicolumn{2}{l}{tenure$^\ddag$} & 1.008 (0.943,1.077) & 0.820 & 0.919\\
\multicolumn{2}{l}{graduate student$^\ddag$} & 0.977 (0.912,1.047) & 0.517 & 0.706\\
\multicolumn{2}{l}{selectivity$^\ddag$} & 0.997 (0.952,1.045) & 0.904 & 0.919\\
\multicolumn{2}{l}{graduation rate$^\ddag$} & 0.978 (0.936,1.022) & 0.320 & 0.623\\
\hline
\end{tabular}}
\resizebox{0.95\textwidth}{!}{\begin{tabular}{l} 
$^\dagger$ The multiplicity-corrected/adjusted p-values were calculated using the method in \cite{benjamini1995controlling}.\\
$^\ddag$ For a numerical variable, the odds ratio is associated with one SD increase.\\
The rows with - as entries are the reference categories for the categorical covariates. The \textbf{bold} rows \\
are the covariates/levels that are statistically significant if $<0.05$ is used for the adjusted p-values.\\
\hline
\end{tabular}}
\end{center}
\end{table}

%------------------------------------------------------

\section{Discussion}\label{sec:discussion}
%what your setting is, why you choose it, its implication and possible modidfication. What other things we did thats not in paper, what are the implications?how your research compare to others. your limitations.

In this study, we collected subReddit data from 128 universities and colleges in the U.S. and some school-level baseline covariates to study sentiment change from 2019 to 2022 that covered the pre-pandemic period to several stages of the COVID-19 pandemic. While we aimed for school representativeness and diversity by considering factors such as school ranking, location, size, and school type, the schools included in this study are not an unbiased sample of all the HEI in the U.S. Since we used subreddit data,  only schools with active subreddits from 2019 to 2022 are eligible, which are the schools that are relatively well-known, larger, and have active online communities on social media. %Also, because our selection process was manual, it reflected the authors' personal impressions of these schools. As a result, our study may not fully represent all U.S. universities and colleges.
For this reason, while the study results can be generalized to the sub-population the data represents and reflects the sentiment changes from 2022 to 2019 in that group, but they would not immediately be generalized to the general population without understanding the demographics of the individuals (which we don't have data on) who posted messages on Reddit.

To reduce labor costs for sentiment labeling, we opted to employ the sentiment classification model proposed in \cite{yan2022covid}, which was trained on data from 2019 and 2020 in eight schools. While there is some overlapping between the training data in \cite{yan2022covid} and the data employed in the current study since they both are extractions from HEI subreddits, the data in this study are much broader and more comprehensive. Therefore, the classifier may not be the most accurate for predicting messages that are outside the range of messages in the training data. %Additionally, the time lag of 1 to 2 years between the training data and our study's data could introduce some bias into the model. 
For future work, we intend to retrain the model using more training data to improve the classification accuracy.

Since the collected subreddit data do not contain individual-level demographic information about the individuals who posted the messages, which can be highly sensitive or pose privacy risks for re-identification, the covariates examined in the GLMM include only school-level public information. The current GLMM does not examine time-varying covariates except for the year itself. A potential interesting extension to the current study is to include time-varying covariates, such as the unemployment rate and the inflation rate, in the model. While the GLMM model itself is not capable of drawing causal relations, it suggests a significant drop in negative sentiment in 2021 compared to 2020, likely due to the availability of vaccines and more effective treatments for COVID-19, giving people hope and a positive outlook that things would return to normal. Though the negative sentiment level in 2022 is still lower than in 2020, it is higher than in 2021, which may be within normal fluctuation in sentiment or indeed reflect a slight rise in negative sentiment from the transient large drop in 2020, due to other factors that negatively affect sentiment such as inflation in 2022. However, this is just speculation and the findings should be regarded as being hypothesis-generating to be confirmed by a rigorous study with a proper set of data that focuses on understanding the reasons behind the emotion shift.

%Our study possesses the potential for extension to encompass other large-scale societal events, enabling the quantification of sentiment changes and the exploration of significant influencing factors. For instance, the extensive layoffs within the tech industry in 2022 have posed challenges in terms of employment opportunities for students majoring in fields like mathematics, statistics, and computer science. To gauge the progress of these layoffs and the onset of recruiting activities, we can consider selecting schools with a technology-focused orientation and analyze their sentiment changes during this specific period. This analysis could serve as an indicator of whether the layoffs are subsiding and recruitment efforts are recommencing.

\section{Conclusion}\label{sec:conclusion}

In this study, we gathered subreddit messages from 128 HEIs in the U.S., covering the pre-pandemic period (2019) to various stages of the pandemic (2020, 2021, and 2022). The sentiments of the messages were predicted using the machine learning procedure in \cite{yan2022covid}.  Adjusting for the school-level covariates, the GLMM analysis suggests a near-full recovery in the sentiment composition (negative vs. non-negative) in 2021 relative to the pre-pandemic era and the negative sentiment levels slightly arose in 2022 but were still notably lower than in 2020.  The results are expected but quantify the sentiment shift from 2019 to 2022. The results also suggest larger enrollment tends to be associated with a higher level of negative sentiment in a statistically significant manner and schools with very high research activities also exhibit more negative sentiments in comparison to schools classified as baccalaureate or Master's colleges/universities. %Our study serves as an illustrative example of harnessing real-world data to unearth factual information pertaining to real-world events, their impacts, and the potential underlying factors contributing to these observations.

\subsection*{Acknowledgments}
Yan is supported by the China Scholarship Council Scholarship. %We thank the editor for taking the time to review our manuscript.
%T.Y. is supported by the China Scholarship Council Scholarship (https://www.chinesescholarshipcouncil.com/).  The funders had no role in study design, data collection and analysis, decision to publish, or preparation of the manuscript.

\subsection*{Appendix}

\subsubsection*{A: List of HEIs included in this study (alphabetically)}
Amherst College, Arizona State University Campus Immersion, Binghamton University, Boise State University, Boston College, Boston University, Bowdoin College, Bowling Green State University-Main Campus, Brigham Young University, Brown University, California Institute Of Technology, California State University-Fullerton, California State University-Northridge, Carleton College, Carnegie Mellon University, Case Western Reserve University, Central Michigan University, Clemson University, Colorado State University-Fort Collins, Columbia University in the City of New York, Cornell University, Dartmouth College, Depaul University, Drexel University, Duke University, Emory University, Florida International University, Florida State University, George Mason University, George Washington University, Georgetown University, Georgia Institute of Technology-Main Campus, Georgia State University, Grinnell College, Harvard University, Illinois Institute of Technology, Indiana University-Bloomington, Iowa State University, Johns Hopkins University, Kansas State University, Marquette University, Michigan State University, Montana State University, New Jersey Institute Of Technology, New York University, North Carolina State University at Raleigh, North Dakota State University-Main Campus, Northwestern University, Ohio State University-Main Campus, Oklahoma State University-Main Campus, Oregon State University, Pomona College, Princeton University, Purdue University-Main Campus, Reed College, Rensselaer Polytechnic Institute, Rice University, Rutgers University-New Brunswick, San Diego State University, Stanford University, Stony Brook University, SUNY at Albany, Syracuse University, Temple University, Texas A \& M University-College Station, Texas Tech University, The College of William and Mary, The Pennsylvania State University, The University of Texas at Austin, Tulane University of Louisiana, United States Naval Academy, University at Buffalo, University of Alabama at Birmingham, University of Arizona, University of Arkansas, University of California-Berkeley, University of California-Los Angeles, University of California-San Diego, University of California-Santa Cruz, University of Chicago, University of Cincinnati-Main Campus, University of Dayton, University of Florida, University of Georgia, University of Hawaii at Manoa, University of Houston, University of Idaho, University of Illinois Chicago, University of Illinois Urbana-Champaign, University of Iowa, University of Kansas, University of Kentucky, University of Louisiana at Lafayette, University of Maine, University of Maryland-College Park, University of Massachusetts-Amherst, University of Memphis, University of Miami, University of Michigan-Ann Arbor, University of Minnesota-Twin Cities, University of Mississippi, University of Missouri-Columbia, University of Nebraska-Lincoln, University of Nevada-Las Vegas, University of New Hampshire-Main Campus, University of New Mexico-Main Campus, University of North Carolina at Chapel Hill, University of North Carolina at Charlotte, University of North Dakota, University of Notre Dame, University of Oklahoma-Norman Campus, University of Pennsylvania, University of Pittsburgh-Pittsburgh Campus, University of South Carolina-Columbia, University of South Florida, University of Utah, University of Vermont, University of Virginia-Main Campus, University of Washington-Seattle Campus, University of Wisconsin-Madison, Vanderbilt University, Villanova University, Virginia Tech, Wake Forest University, Washington University in St Louis, West Virginia University, Western Michigan University, Yale University.

\subsubsection*{B: Data}
The data we collected from Reddit for this study include 1) userID,  2) time of a posted message, 3) content of a message, 4) dyadic relation between two messages in terms of whether one is a  reply to the other. There is no information in the collected data that allows direct identification of a user in the data.  The information in 'userID' was not used in our model training or methodological development and was replaced by dummy integers  after the data were downloaded.  

The data collection is in compliance with the Reddit privacy policy \cite{redditprivacy}, Reddit user agreement \cite{reddituser},  and Reddit API’s term of use \cite{redditapi}. The data are anonymized in the sense that we replaced the actual Reddit userIDs in the data with dummy IDs (integers  1, 2, ...), further limiting potential privacy risk (e.g., through record linkage) without losing information for learning tasks based on the data. The unprocessed data with actual Reddit userIDs replaced by dummy IDs are available at 
\url{https://drive.google.com/file/d/1p4ZDeg2UWPw6uZwRYsnsHa9QhXWFVODK/view?usp=sharing} and \url{https://drive.google.com/file/d/19pCRwE0AyGkHBC97CMghjqPC8PZhX0PG/view?usp=sharing}, and the RoBERTa-processed data and scores are available at 
\url{https://drive.google.com/file/d/1YXt2z4sGJABRvdTwQLgAt6wIZvH4y_w2/view?usp=sharing}.

\subsubsection*{C: Code}
The code for this study can be downloaded from \url{https://github.com/AlvaYan/postCOVIDSentiAnalysis}). The  Python code for  the  RoBERTa framework that we applied  is adapted from \cite{barbieri2020tweeteval} and is available at \url{https://huggingface.co/cardiffnlp/twitter-roberta-base-sentiment};  the Python code for training the GAT NN is adapted from \cite{wang2019heterogeneous} (\url{https://github.com/Jhy1993/HAN}).
% and available at \url{https://huggingface.co/cardiffnlp/twitter-roberta-base-sentiment} 

%\bibliography{CIT}

\end{document}